\documentclass[conference]{IEEEtran}
\IEEEoverridecommandlockouts
\usepackage{cite}
\usepackage{amsmath,amssymb,amsfonts}
\usepackage{algorithmic}
\usepackage{graphicx}
\usepackage{textcomp}
\usepackage{xcolor}
\usepackage{mathrsfs}
\usepackage{gensymb}  
\usepackage{array}
\usepackage{url}
\usepackage{dblfloatfix}
\usepackage{caption}
\usepackage{enumerate}
\usepackage{booktabs}
\usepackage{adjustbox}
\usepackage{multirow}
\newtheorem{remark}{Remark}
\usepackage{fancyhdr} 

\def\BibTeX{{\rm B\kern-.05em{\sc i\kern-.025em b}\kern-.08em
    T\kern-.1667em\lower.7ex\hbox{E}\kern-.125emX}}
\begin{document}

\title {Integration of Mamba and Transformer -- MAT for Long-Short Range Time Series Forecasting with Application to Weath Dynamics}

\author{\IEEEauthorblockN{1\textsuperscript{st} Wenqing Zhang}
\IEEEauthorblockA{\textit{McKelvey School of Engineering} \\
\textit{Washington University in St. Louis}\\
St. Louis, USA \\
wenqing.zhang@wustl.edu}
\and
\IEEEauthorblockN{2\textsuperscript{nd} Junming Huang}
\IEEEauthorblockA{\textit{Heinz College} \\
\textit{Carnegie Mellon University}\\
Kirkland, USA \\
junmingh14@gmail.com}
\and
\IEEEauthorblockN{3\textsuperscript{rd} Ruotong Wang}
\IEEEauthorblockA{\textit{Department of Computer Science} \\
\textit{University at Albany, State University of New York}\\
Albany, USA \\
rwang31@albany.edu}
\and
\IEEEauthorblockN{4\textsuperscript{th} Changsong Wei}
\IEEEauthorblockA{\textit{Independent Researcher} \\
\textit{Digital Financial Information Technology Co.LTD}\\
Chengdu, China \\
changsongwei88@gmail.com}
\and
\IEEEauthorblockN{5\textsuperscript{th} Wenqian Huang}
\IEEEauthorblockA{\textit{Carey Business School} \\
\textit{Johns Hopkins University}\\
Baltimore, USA \\
crystalhuang1031@gmail.com}
\and
\IEEEauthorblockN{6\textsuperscript{th} Yuxin Qiao*}
\IEEEauthorblockA{\textit{Department of  Information Technology} \\
\textit{Northern Arizona University}\\
Flagstaff, USA \\
yq83@nau.edu}
}

\maketitle  
\thispagestyle{fancy}            
\fancyhead{}                     
\rhead{Proc. of the 5th International Conference on Electrical, Communication and Computer Engineering (ICECCE)} 

\begin{abstract}
Long-short range time series forecasting is essential for predicting future trends and patterns over extended periods. While deep learning models such as Transformers have made significant strides in advancing time series forecasting, they often encounter difficulties in capturing long-term dependencies and effectively managing sparse semantic features. The state space model, Mamba, addresses these issues through its adept handling of selective input and parallel computing, striking a balance between computational efficiency and prediction accuracy. This article examines the advantages and disadvantages of both Mamba and Transformer models, and introduces a combined approach, MAT, which leverages the strengths of each model to capture unique long-short range dependencies and inherent evolutionary patterns in multivariate time series. Specifically, MAT harnesses the long-range dependency capabilities of Mamba and the short-range characteristics of Transformers. Experimental results on benchmark weather datasets demonstrate that MAT outperforms existing comparable methods in terms of prediction accuracy, scalability, and memory efficiency.
\end{abstract}


\section{Introduction}
Long-short range time series forecasting (LSRTSF) has demonstrated indispensable value across various modern domains, including demand prediction, signal processing and detection, and regulation of critical assets such as renewable energy, greenhouse farming, and advanced manufacturing facilities~\cite{torres2021deep,guo2024hybrid}. Although several practical techniques for LSRTSF have been proposed in recent years, most of these techniques primarily focus on a single aspect. These aspects include handling long-term dependencies in complex scenarios, achieving linear scalability in model design, or enhancing computational efficiency and integration with edge computing. However, these techniques still face numerous challenges posed by the big data era and the need for better long-term forecasting capabilities.

Effectively capturing long-term dependencies in multivariate time-series (MTS) data is paramount for accurate and robust forecasting. Linear models like DLinear and TiDE have demonstrated the capability to achieve performance comparable to that of Transformer-based models while maintaining low complexity and excellent scalability\cite{zeng2023transformers,das2023long}. These models primarily employ multilayer perceptrons and linear projection techniques, which, although efficient, may sometimes fall short in capturing subtle long-range correlations inherent in the data. On the other hand, Transformer-based models, including iTransformer, PatchTST, and Crossformer, excel in uncovering long-distance dependencies due to their sophisticated self-attention mechanisms. This capability significantly enhances the accuracy of LSRTSF. Nevertheless, despite their advanced features and superior performance, Transformer-based models often grapple with challenges related to scalability and practicality, especially in edge computing environments, due to their high computational complexity\cite{han2022survey,ni2024timeseries}. This trade-off between accuracy and computational efficiency remains a critical consideration in the development and deployment of MTS forecasting models\cite{liu2023survey}.

Recently, state-space models (SSMs) have garnered significant attention due to their exceptional performance in sequence data inference\cite{rangapuram2018deep}. These models handle very long sequences with linear complexity and demonstrate context-aware selectivity through hidden attention mechanisms\cite{rangapuram2018deep}. While SSMs have shown potential in various domains such as genomics, table learning, graph data, and imagery, their application to LSRTSF remains underexplored\cite{patro2024mamba}. The primary reasons for this are the recent development of SSM techniques with strong content and context selectivity and the ongoing challenge of efficiently representing contextual information in time series data\cite{lieber2024jamba}. Transformer-based models like Autoformer and Informer treat each time point as a separate unit, while newer models such as PatchTST and iTransformer process segments of the time series\cite{patro2024mamba}.

MTS data typically comprises multiple channels, each representing a different variable. Models like Informer, FEDformer, and Autoformer utilize a mixed-channel approach, treating the input as a two-dimensional matrix defined by the number of channels and the length of histories\cite{xu2023multimodal}. This approach is beneficial when channels exhibit significant correlations, as it allows for the capture of these dependencies effectively. The temporal relationships inherent in time series data, which tend to be preserved even after downsampling, have been explored in models like SciNet but remain underutilized in many other approaches\cite{liu2022scinet}. Due to the high redundancy in MTS data, relying solely on time points as markers can obscure context-based choices and overlook long-range dependencies. Instead, utilizing data snippets within a time window can provide richer contextual clues. To optimally capture long-range dependencies, it is crucial to provide context at multiple scales. This can be achieved by automatically generating global-level markers as context, similar to the approach used by iTransformer, which models the entire recall window to enhance understanding and prediction accuracy\cite{liu2023itransformer}.

This study introduces a novel method designed to efficiently capture long- and short-range dependencies in time-series data by providing multi-scale contexts, with a particular focus on enhancing local contexts. The approach employs a selectively scanning SSM known as Mamba, coupled with attention mechanisms called Transformer to serve as the core inference engine, termed MAT. This architecture enables the capture of both long-term forecasting capabilities and short-range dependencies in MTS data while maintaining linear scalability and minimal memory usage. The model leverages the distinct characteristics of time-series data by employing a bottom-up strategy, which involves generating contextual cues at two distinct scales through linear mapping for resolution reduction or downsampling. The high-resolution level functions at the first scale, whereas the low-resolution level operates at the second scale. At each of these levels, the model utilizes four MAT modules: one for collecting local context clues and another for gathering both global context clues, which are then fused.

The main contribution of the proposed algorithm could be concluded as:
\begin{itemize}
	\item This research introduces an innovative model, MAT, representing a pioneering step in capturing both long-term and short-range dependencies in multivariate time-series data through exclusive use of MAT modules for context-aware prediction. MAT showcases linear scalability and a minimal memory footprint, delivering performance that is either superior or comparable to existing linear models.
	
	\item The architecture of MAT is uniquely designed to uniformly handle variable sequence lengths and extract the long-short range relationships within the time series. It leverages inter-channel correlations using four MAT modules, integrating with Transformer and Mamba. Furthermore, MAT efficiently selects predictive content by targeting global and local contextual information at various scales of MTS data, thereby optimizing its ability to extract meaningful long-short range patterns from complex observations.
\end{itemize}

\section{Related Work}
Various methods for the ubiquitous LSRTSF problem have been developed, falling into three primary categories regarding this article: Self-supervised methods, Transformer, and Mamba.

\textbf{Self-Supervised Methods:}  Traditional solutions rely on statistical properties and patterns in data for prediction. Recently, there has been a shift towards deep learning approaches, which employ neural networks to capture complex dependencies and improve performance. These methods are categorized into variable-mixing and variable-independent approaches. The first branch, including RNNs, LSTM, and GRU, model dependencies across time and variables, despite challenges in optimization efficiency\cite{koutnik2014clockwork,yu2019review,yamak2019comparison}. Variable-independent methods assume variable independence, offering simplicity and efficiency but potentially oversimplifying the problem\cite{zhou2023one}.

\textbf{Transformer:} Transformer-based models have become widely adopted in LSRTSF for their remarkable accuracy\cite{vaswani2017attention}. Examples include iTransformer, PatchTST, Crossformer, FEDformer, Stationary, and Autoformer\cite{lin2022survey}. These models transform time series data into token sequences and utilize a self-attention mechanism to extract dependencies across various time steps, making them particularly adept at capturing complex temporal relationships. Additionally, Transformers enjoy the ability to process data in parallel in order to enhance their effectiveness in identifying long-term dependencies, sometimes with linear scalability. Despite their advantages, these models often struggle with quadratic time and memory complexity due to the nature of self-attention mechanisms\cite{xu2023multimodal}.

\textbf{State Space Models:} Recently, some studies have integrated SSMs with deep learning, showcasing substantial potential in addressing long-term dependency issues~\cite{rangapuram2018deep}. However, their computational and memory demands often hinder practical applications~\cite{gu2021efficiently}. Efficient variants of SSMs, such as S4, H3, Gated State Space, and RWKV\cite{zhang2024survey}, aim to improve performance and efficiency in diverse tasks. Mamba\cite{gu2023mamba} addresses a major limitation of traditional SSMs by introducing an S4-based data-dependent selection technique, effectively filtering specific inputs and capturing long-term context as the sequence length extends. Mamba achieves linear time efficiency in modeling long sequences and outperforms Transformer models in benchmark tests~\cite{gu2023mamba}. Mamba has been effectively employed across various domains, including audio signals, textual data, sensor readings, and genomics sequences\cite{patro2024mamba}. This extensive application has greatly enhanced its capability to detect intricate patterns and long range dependencies in diverse types of data. In addition, a notable research direction involves integrating the Transformer and Mamba for language modeling\cite{park2024can}. Comparative studies\cite{park2024can} demonstrate that the Mambaformer is highly effective for in-context learning tasks. Jamba\cite{lieber2024jamba}, the first production-grade attention-SSM hybrid model, boasts 12 billion active and 52 billion total available parameters, exhibiting excellent performance with long-context tasks.

\section{Methodology}
\subsection{Preliminary}
In this section, we provide a comprehensive overview of each component of our proposed architecture and elucidate how our model tackles the LSRTSF problem. Consider a collection of MTS samples, denoted as dataset $\mathcal{D} $. Each sample consists of an input sequence $ \mathbf{x} = [x_1, \ldots, x_L]$, where each $x_t \in \mathbb{R}^M$ represents a vector of $M$ measurements at time point $t$. The sequence length $L$, also referred to as the look-back window, serves as the basis for predicting $T$ future values, denoted by $ [x_{L+1}, \ldots, x_{L+T}]$.

\subsection{Transformer}
\begin{figure}[h]
	\centering
		\includegraphics[width=0.25\textwidth]{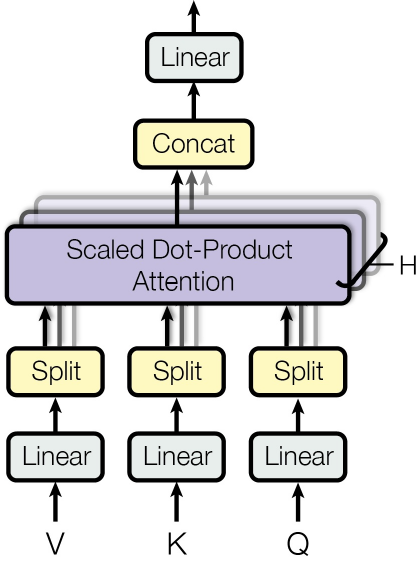}
	\caption{Visualization of the Transformer Mechanism.}
	\label{f:fig1}
\end{figure}

At the core of the proposed algorithm, as well as other transformer-based approaches, lies the concept of attention, which enables the models to concentrate on significant elements within a context, which has been visualized in Fgiure.\ref{f:fig1}. One example is multi-head attention, which transforms a sequence of queries $Q \in \mathbb{R}^{\ell_q \times d}$ of length $\ell_q$ into a sequence of outputs $O = [O_1, \ldots, O_H] \in \mathbb{R}^{\ell_q \times d}$ of the same size by attending to $\ell_k$ key-value pairs $K \in \mathbb{R}^{\ell_k \times d}$, $V \in \mathbb{R}^{\ell_k \times d}$:
\begin{equation}
	O_h = \text{Attention}(Q_h, K_h, V_h) = \text{Softmax}\left(\frac{Q_h K_h^T}{\sqrt{d}}\right) V_h,
\end{equation}
where $Q_h = QW_Q^h$, $K_h = KW_K^h$, $V_h = VW_V^h$ are the projected queries, keys, and values for head $h \in [1, H]$ with learning parameters $W_Q^h$, $W_K^h$, and $W_V^h$, respectively. When $Q = K = V$, this attention mechanism is referred to as self-attention.

Given fully observed input sequences, the mapping can be computed efficiently without the sequential order constraints typically imposed by recurrent neural networks. Crucially, the mechanism inherently includes direct connections between distant time steps, allowing information from previous time steps to be accessed without compression into a fixed representation. This simplifies the optimization and learning of long-term dependencies\cite{vaswani2017attention,li2024lightgbm}.

Without recurrence, the Transformer model\cite{vaswani2017attention} encodes information about each time step $t$ using predefined sinusoidal positional embeddings $\text{Position}(t) = [p_t(1), \ldots, p_t(d)] \in \mathbb{R}^d$, where the $i$-th embedding is defined as $p_t(i) = \sin(t \cdot c^{i/d})$ for even $i$ and $p_t(i) = \cos(t \cdot c^{i/d})$ for odd $i$, with $c$ being a large constant.

\subsection{Mamba}
\begin{figure}[h]
	\centering
	\includegraphics[width=0.4\textwidth]{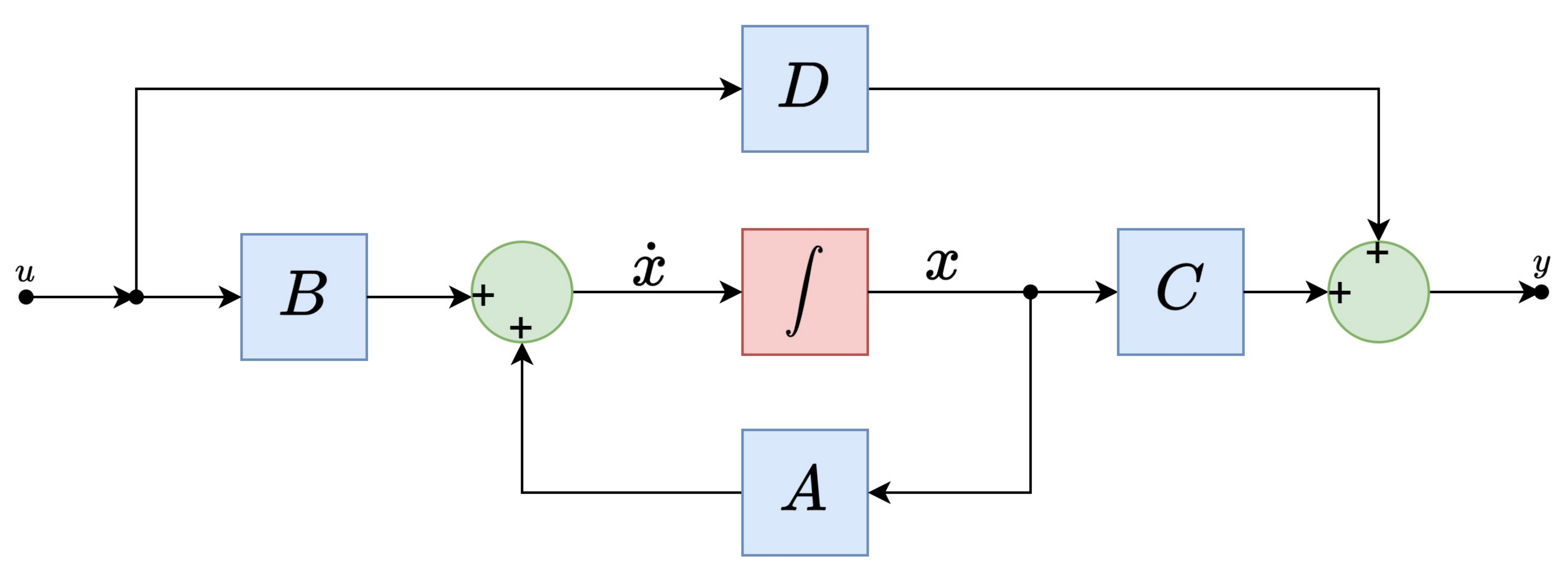}
	\caption{Illustration demonstrating the concept of the SSM.}
	\label{f:fig2}
\end{figure}

\textbf{SSMs} are commonly conceptualized as linear time-invariant (LTI) systems that transform continuous input signals $ u(t) $ into corresponding output signals $ y(t) $ via a system state representation $ x(t) $. This state space framework, which has been presented as Figure.\ref{f:fig2}, delineates the temporal evolution of the state, which can be described by the following set of ordinary differential equations:
\begin{align}
	\dot{x}(t) &= A x(t) + B u(t) \\ \notag
	y(t) &= C x(t) + D u(t)
\end{align}
where $ \dot{x}(t) = \frac{d x(t)}{dt} $, and $ A, B, C, $ and $ D $ are the parameters of the time-invariant SSMs.

\begin{figure*}
	\centering
	\includegraphics[width=\textwidth,height= 0.35\textwidth]{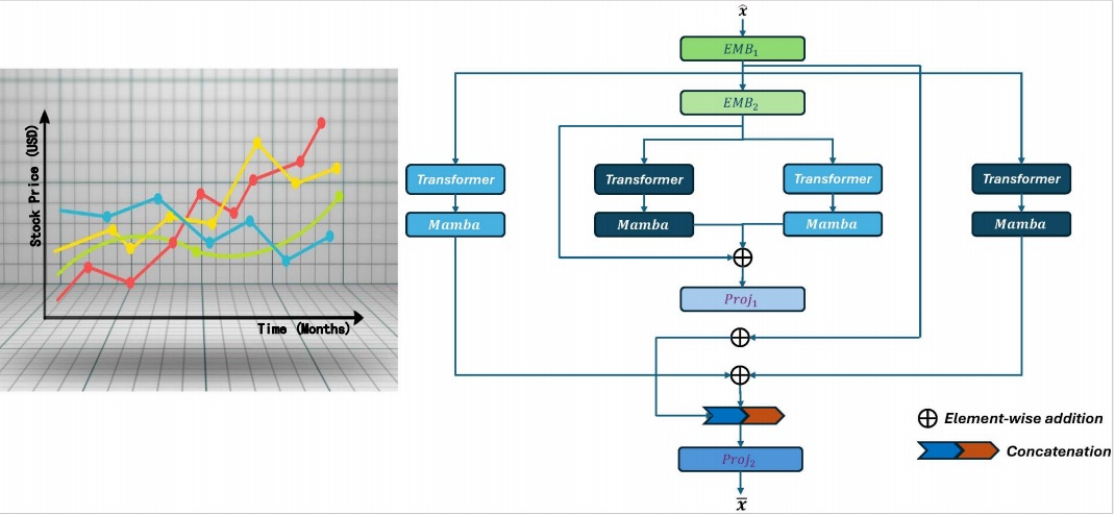}
	\caption{Workflow of the Proposed MAT.}
	\label{f:fig4}
\end{figure*}

\textbf{Discretization}: Solving SSMs analytically is exceedingly difficult due to their continuous nature. To address this, discretization techniques are employed to approximate the continuous-time SSM into a discrete-time counterpart. This involves sampling the input signals at fixed intervals, yielding their discrete-time equivalents. The resultant discrete-time SSM is expressed as:
\begin{align}
	x_k &= \bar{A} x_{k-1} + \bar{B} u_k \\ \notag
	y_k &= \bar{C} x_k + \bar{D} u_k
\end{align}
where $ x_k $ denotes the state vector at discrete time step $ k $, and $ u_k $ denotes the input vector at the same step. The matrices $ A $ and $ B $ in the discrete domain are derived from their continuous-time counterparts using discretization techniques like the Euler method or the Zero-Order Hold (ZOH) method. Specifically, $ \bar{A} = \exp(\varDelta A) $ and $ \bar{B} = (\varDelta A)^{-1}(\exp(\varDelta A) - I)\varDelta B $, $\varDelta$ is the sampling time interval.

\begin{figure}[h]
	\centering
	\includegraphics[width=0.25\textwidth, height=0.3\textwidth]{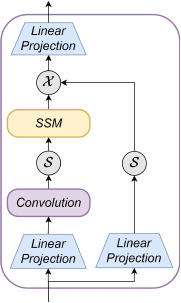}
	\caption{Mamba using the SSM.}
	\label{f:fig3}
\end{figure}

\textbf{Selective Scan Mechanism}: Mamba enhances traditional SSMs by incorporating a selective mechanism, allowing parameters to modulate interactions along the sequence contextually, which has been demonstrated by Figure.\ref{f:fig3}. This selective approach enables Mamba to filter out irrelevant noise in time series tasks and selectively retain or discard information pertinent to the current input. Unlike previous SSM methodologies with static parameters, this approach deviates from the LTI characteristics. Consequently, Mamba employs a hardware optimization strategy and implements parallel scan training to effectively address this challenge.

\begin{table*}[htbp]
	\centering
	\begin{adjustbox}{width=\textwidth}
		\begin{tabular}{cccccccccccccc}
			\toprule
			\multicolumn{2}{c}{Methods} & \multicolumn{2}{c}{TimeMachine} & \multicolumn{2}{c}{iTransformer} & \multicolumn{2}{c}{RLlinear} & \multicolumn{2}{c}{PatchTST} & \multicolumn{2}{c}{Crossformer} & \multicolumn{2}{c}{Autoformer} \\
			\cmidrule(lr){3-4} \cmidrule(lr){5-6} \cmidrule(lr){7-8} \cmidrule(lr){9-10} \cmidrule(lr){11-12}\cmidrule(lr){13-14}
			$\mathcal{D}$ & $\mathcal{T}$ & MSE & MAE & MSE & MAE & MSE & MAE & MSE & MAE & MSE & MAE & MSE & MAE \\
			\midrule
			\multirow{4}{*}{Weather} & 96  & \underline{0.167} & \textbf{0.211} & 0.174 & \underline{0.214} & 0.192 & 0.232 & 0.177 & 0.218 & \textbf{0.158} & 0.230 & 0.266 & 0.336 \\
			& 192 & \underline{0.213}& \textbf{0.292} & 0.221 & \underline{0.254} & 0.240 & 0.271 & 0.225 & 0.259 & \textbf{0.206} & 0.277 & 0.307 & 0.367  \\
			& 336 & \textbf{0.257} & \textbf{0.293} & 0.278 & \underline{0.296} & 0.292 & 0.307 & 0.278 & 0.297 & \underline{0.272} & 0.335 & 0.359 & 0.395 \\
			& 720 & \textbf{0.347} & \textbf{0.348} & 0.358 & \underline{0.349} & 0.364 & 0.353 & 0.354 & 0.348 & 0.398 & 0.418 & 0.419 & 0.428 \\
			\bottomrule
		\end{tabular}
	\end{adjustbox}
	\begin{adjustbox}{width=\textwidth}
		\vspace{50pt}
		\begin{tabular}{cccccccccccccc}
			\toprule
			\multicolumn{2}{c}{Methods} & \multicolumn{2}{c}{TiDE} & \multicolumn{2}{c}{TimesNet} & \multicolumn{2}{c}{DLinear} & \multicolumn{2}{c}{SCINet} & \multicolumn{2}{c}{FEDformer} & \multicolumn{2}{c}{Stationary} \\
			\cmidrule(lr){3-4} \cmidrule(lr){5-6} \cmidrule(lr){7-8} \cmidrule(lr){9-10} \cmidrule(lr){11-12} \cmidrule(lr){13-14} 
			$\mathcal{D}$ & $\mathcal{T}$ & MSE & MAE & MSE & MAE & MSE & MAE & MSE & MAE & MSE & MAE & MSE & MAE \\
			\midrule
			\multirow{4}{*}{Weather} & 96  & 0.202 & 0.261 & 0.172 & 0.220 & 0.196 & 0.255 & 0.221 & 0.306 & 0.217 & 0.296 & 0.173 & 0.223 \\
			& 192 & 0.242 & 0.298 & 0.219 & 0.261 & 0.237 & 0.296 & 0.261 & 0.340 & 0.276 & 0.336 & 0.245 & 0.285 \\
			& 336 & 0.287 & 0.306 & 0.280 & 0.306 & 0.283 & 0.335 & 0.309 & 0.378 & 0.339 & 0.380 & 0.321 & 0.338 \\
			& 720 & \underline{0.351}& 0.386 & 0.365 & 0.359 & 0.345 & 0.381 & 0.377 & 0.427 & 0.403 & 0.428 & 0.414 & 0.410 \\
			\bottomrule
		\end{tabular}
	\end{adjustbox}
	\caption{Performance verification of Weather datasets. The best results are in \textbf{bold} and the second best are \underline{underlined}.}
	\label{tab:weather}
\end{table*}

\subsection{MAT}
The framework of the proposed model, named MAT, is illustrated in Figure. \ref{f:fig4}. The core components of this algorithm are four combined Mambas and Transformer, which are employed to extract long-short range contextual information. It is noted that this configuration leverages both the long-term forecasting capability of Mamba and short range dependency of Transformer.

Before streaming the collected datasets into our procedure, the original MTS is normalized into $\mathbf{\hat{x}} = [\hat{x}_{1}, \cdots, \hat{x}_{L}] \in \mathbb{R}^{M \times L}$ using $\mathbf{\hat{x}} = \text{Normalize}(\mathbf{x})$. This normalization operation is processed by the reversible instance normalization (RevIN), which is widely utilized in MTS-related problems\cite{kim2021reversible,ma2024fostc3net}. Furthermore, the identical normalization process, e.g. RevIN, has been implemented in the data-preprocessing module of the baseline algorithm in the experiments.

To reduce the dimensionality of the original data and accelerate subsequent inference of MAT,  a two-stage embedding representation technique of the input is employed as:
\begin{equation}
\mathbf{x}^{(1)} = EMB_1(\mathbf{x}^{(0)}),  \mathbf{x}^{(2)} = EMB_2(\mathbf{x}^{(1)})
\end{equation}
where the embedding functions $EMB_1: \mathbb{R}^{M \times L} \rightarrow \mathbb{R}^{M \times n_1}$ and $EMB_2: \mathbb{R}^{M \times n1} \rightarrow \mathbb{R}^{M \times n_2}$ are implemented through multi-layer perceptrons (MLPs). In addition, the dropout operation is introduced in the calculation of the embedding function $E_2$ to mitigate overfitting phenomenon. To deal with the variable input sequence length $L$, a fixed-length parameters, which are $n_1$ and $n_2$, are chosen from the set $\{512, 256, 128, 64, 32\}$ such that $n_1 > n_2$.

To be specific, the proposed algorithm has a similar procedure with the Mamba-based method in\cite{ahamed2024timemachine}, there exists a main objective-level difference, the MAT approach in this paper aims at not only the long-term prediction capability of the prevalent Mamba methods, but also the short-range dependency learned from the Transformer module, which has been proved to be a pivotal component in the modern large language model \cite{ni2024earnings,song2023zeroprompt,jin2024apeer}. This enhanced phenomenon has been observed and analyzed in the pioneering work in\cite{xu2024integrating}, where the reader could refer to more details. To be specific, the MTS problem could be decomposed into two main branches, one is the long term and another is short range, Mamba uses the long-proved SSMs to overcome the long-term prediction dilemma and Transformer uses the core attention mechanism to strengthen the short range dependent relationship. The proposed MAT enjoys the long-short prediction capability of them and outperforms many existing algorithms in this area.
\begin{remark}
As highlighted in the pioneering work \cite{xu2024integrating}, the Mamba and Transformer architectures have significantly advanced in achieving superior performance in MTS scenarios. However, effectively and efficiently integrating their distinct advantages remains an open problem. In this paper, the authors make a slight modification to the traditional method, resulting in notable enhancements. Future research should focus on deeply investigating and interchangeably fusing the latest developments in Mamba and Transformer methodologies\cite{wang2024token,sharma2024focating,friedman2024learning}.
\end{remark}

To output the predicted time sequence of the identical length with the original MTS from MAT, the corresponding two MLPs, $Proj_1$ and $Proj_2$, which produce $n _1$ and $T$ time points respectively. To be specific, the projector $Proj_1$ map the output $\mathbb{R}^{M \times n_2} \rightarrow \mathbb{R}^{M \times n_1}$ to obtained $\mathbf{\bar{x}}^1$, the subsequent projector $Proj_2$ map the above-mentioned result 
$\mathbb{R}^{M \times n_1} \rightarrow \mathbb{R}^{M \times T}$ to yield $\mathbf{\bar{x}}$. The mathematical formulation of this proceeding process could be summarized as 
\begin{equation}
	\mathbf{\bar{x}}^{(1)} = Proj_1(\mathbf{\bar{x}}^{(F_1)}),  \mathbf{\bar{x}} = Proj_2(\mathbf{\bar{x}}^{(1)}\oplus \mathbf{\hat{x}}^{(F_2)})
\end{equation}

It should be noted that this two-stage projection protocol is systematic with the two-stage embedding operation in $EMB_1$ and $EMB_2$ and the residual connections are widely utilized in the developed methodology to avoid the zero-gradient problem in the deep learning topic and enrich the learned feature in every layer. 

\section{Experiments}
This section showcases the primary outcomes of our experiments using well-established weather benchmark datasets for MTS forecasting. Furthermore, we perform a comprehensive analysis to highlight the contribution of the proposed methodology.

\subsection{Dataset}
The weather datasets, \url{https://www.bgc-jena.mpg.de/wetter/}, recorded every 10 minutes throughout the entire year of 2020, encompass 21 meteorological indicators, including air temperature and humidity. Provided by the Max Planck Institute for Biogeochemistry, this dataset includes detailed, long-term time series data on temperature, precipitation, and real-time weather conditions specifically for Jena, Germany. It features comprehensive statistical analyses, information on sun and moon phases, and integrates data from external weather stations and webcams. For a fair comparison, the results verified by the work\cite{ahamed2024timemachine} are used in this paper.

\subsection{Setting}
All experiments were performed using the PyTorch framework on one NVIDIA V100 GPU with 32GB of memory. The model optimization was carried out using the ADAM algorithm, with $L_2$ loss as the objective function and the learning rate $\eta = 0.0001$. The batch size was adjusted according to the prediction requirement, and the training duration was set to 100 epochs. Prediction accuracy was evaluated using mean square error (MSE) and mean absolute error (MAE) metrics, with lower values signifying higher accuracy.

\subsection{Evaluation}
Table.\ref{tab:weather} presents the results of MAT on supervised long-short range forecasting tasks. Adhering to the widely adopted settings, all baselines, including the MAT in this article, were fixed with $L = 96$ and $T = \{96, 192, 336, 720\}$. Additionally, default parameters for all Mambas were set as follows: Dimension factor $D = 256$, local convolutional width = 2, and state expand factor $N = 1$, for the transformer module, the multi head number is set $H = 8$, the Batch Size in the training process is set as $Batch = 32$.

The results in Table.\ref{tab:weather} indicate that the proposed MAT outperforms many published baselines across weather datasets. Notably, Crossformer achieves superior results on these weather datasets, which have numerous channels and complicated neural network construction. Our method, however, shows comparable or superior performance on weather datasets, significantly surpassing the existing baselines. This underscores the effectiveness of the developed MAT in dealing with LSRTSF tasks across various scales and complexity.

\subsection{Analysis}
Table.\ref{tab:weather} demonstrates the effectiveness of MAT using MAE and MSE metrics on the weather datasets. It is evident that MAT accurately captures real-world trends in the predicted future time horizon for the testing samples. Specifically, on the weather datasets, MAT shows superior performance compared to published prediction algorithms. However, it is notable that MAT was outperformed by Crossformer in shorter sequence lengths. A potential explanation for this behavior could be the conflict of interest between the long-term prediction capabilities of Mamba and the short-range dependency strengths of Transformer. Future research could focus on seamlessly integrating Mamba and Transformer to address this issue.

\section{Conclusion}
This study addresses the challenges in long-short range time series forecasting by comparing Transformers and the SSM, Mamba. While Transformers struggle with long-term dependencies and sparse semantic features, Mamba excels through selective input handling and parallel computing. We introduced MAT, a combined approach leveraging Mamba's long-range capabilities and Transformers' short-range strengths. Experiments on benchmark weather datasets demonstrate that MAT outperforms existing methods in prediction accuracy, scalability, and memory efficiency, making it a promising solution for forecasting in multivariate time series.

\end{document}